\documentclass[letterpaper]{article} 
\usepackage{aaai25}  
\usepackage{times}  
\usepackage{helvet}  
\usepackage{courier}  
\usepackage[hyphens]{url}  
\usepackage{graphicx} 
\usepackage{natbib}  
\usepackage{caption} 
\usepackage{algorithm}

\usepackage{amssymb}
\usepackage{amsmath}
\usepackage{cases}
\usepackage{algpseudocode}
\usepackage{amsmath}
\usepackage{graphicx}
\usepackage{caption}
\usepackage{subcaption}
\usepackage{booktabs}
\usepackage{graphicx}
\usepackage{array}
\usepackage{newtxtext,newtxmath}
\usepackage{multirow}
\usepackage{caption}
\usepackage{graphicx} 
\usepackage{pifont}
\usepackage{amsmath}
\usepackage{makecell}

\usepackage{newfloat}
\usepackage{listings}
\DeclareCaptionStyle{ruled}{labelfont=normalfont,labelsep=colon,strut=off} 
\floatstyle{ruled}
\newfloat{listing}{tb}{lst}{}
\floatname{listing}{Listing}
\title{What Kind of Visual Tokens Do We Need? Training-free Visual Token Pruning for Multi-modal Large Language Models from the Perspective of Graph}
\author{
    Yutao Jiang\equalcontrib\textsuperscript{\rm 1},
    Qiong Wu\equalcontrib\textsuperscript{\rm 1\rm 2},
    Wenhao Lin\textsuperscript{\rm 1\rm 2},
    Wei Yu\textsuperscript{\rm 1\rm 2},
    Yiyi Zhou\thanks{Corresponding author}\textsuperscript{\rm 1\rm 2}
}
\affiliations{
    \textsuperscript{\rm 1}Key Laboratory of Multimedia Trusted Perception and Efficient Computing, \\
    Ministry of Education of China, Xiamen University, 361005, P.R. China.\\

    \textsuperscript{\rm 2}Institute of Artificial Intelligence, Xiamen University, 361005, P.R. China.\\

    \{yutaojiang, qiong, wenhaolin, weiyu\}@stu.xmu.edu.cn, zhouyiyi@xmu.edu.cn

}

\usepackage{bibentry}

\begin{document}

\maketitle

\begin{abstract}
Recent \emph{Multimodal Large Language Models} (MLLMs) often use a large number of visual tokens to compensate their visual shortcoming, leading to excessive computation and obvious visual redundancy.
In this paper, we investigate what kind of visual tokens are needed for MLLMs, and reveal that both foreground and background tokens are critical for MLLMs given the varying difficulties of examples. Based on this observation, we propose a \emph{graph}-based method towards training-free visual token pruning, termed \emph{G-Prune}.
In particular, G-Prune regards visual tokens as nodes, and construct their connections based on their semantic similarities. Afterwards, the information flow is propagated via weighted links, and the most important tokens after iterations are kept for MLLMs, which can be front or background.
To validate G-Prune, we apply it to a recent MLLM called LLaVA-NeXT, and conduct extensive experiments on a set of benchmarks.
The experiment results show that G-Prune can greatly reduce computation overhead while retaining high performance on both coarse- and fine-grained tasks.  
For instance, G-Prune can reduce 63.57\% FLOPs of LLaVA-NeXT on VQA2.0 and TextVQA with only 0.95\% and 2.34\% accuracy drops, respectively.
 Our code is available at \url{https://github.com/jytmelon/G-Prune}.
\end{abstract}

\section{Introduction}
Recently, extending \emph{large language models} (LLMs) to more modalities has become a research hotspot~\cite{achiam2023gpt,bai2023qwen,touvron2023llama2,chen2024internvl,guo2024deepseek}, \emph{i.e.}, multimodal LLM (MLLMs). For vision-language (VL) tasks, the most common paradigm is to directly project the extracted visual features into the semantic space of LLMs \cite{li2023blip,team2023gemini,achiam2023gpt,liu2024visual,liu2024LLaVA,luo2024cheap} as the input tokens. Despite effective, these MLLMs \cite{team2023gemini,achiam2023gpt,liu2024visual,liu2024LLaVA} often suffer from more severe visual hallucinations, and perform worse on granular VL tasks like TextVQA~\cite{singh2019towards}.

A straightforward solution to remedy the visual shortcoming of MLLMs is to increase image resolution, \emph{i.e.}, using more visual tokens, thereby making vision matters in multimodal reasoning. For instance, LLaVA-NeXT~\cite{liu2024LLaVA} partitions images into multiple regions and directly concatenates the visual tokens mapped from each region. InternVL~\cite{chen2024internvl} presets a variety of high-resolution image inputs with different aspect ratios. Although simple, this solution does improve the visual reasoning of MLLMs to a large extent, achieving new SOTA performances on a set of benchmarks. However, a counterpart is the prohibitively expensive computation. For instance, LLaVA-NeXT uses 2880 visual tokens, and require about 18.52 TFLOPs computational for each inference. 

\begin{figure}[tbp]
    \begin{minipage}[c]{0.05\columnwidth} 
        \caption*{(a)} 
        \label{fig:a}
    \end{minipage}%
    \begin{minipage}[c]{0.45\columnwidth} 
        \includegraphics[width=\linewidth]{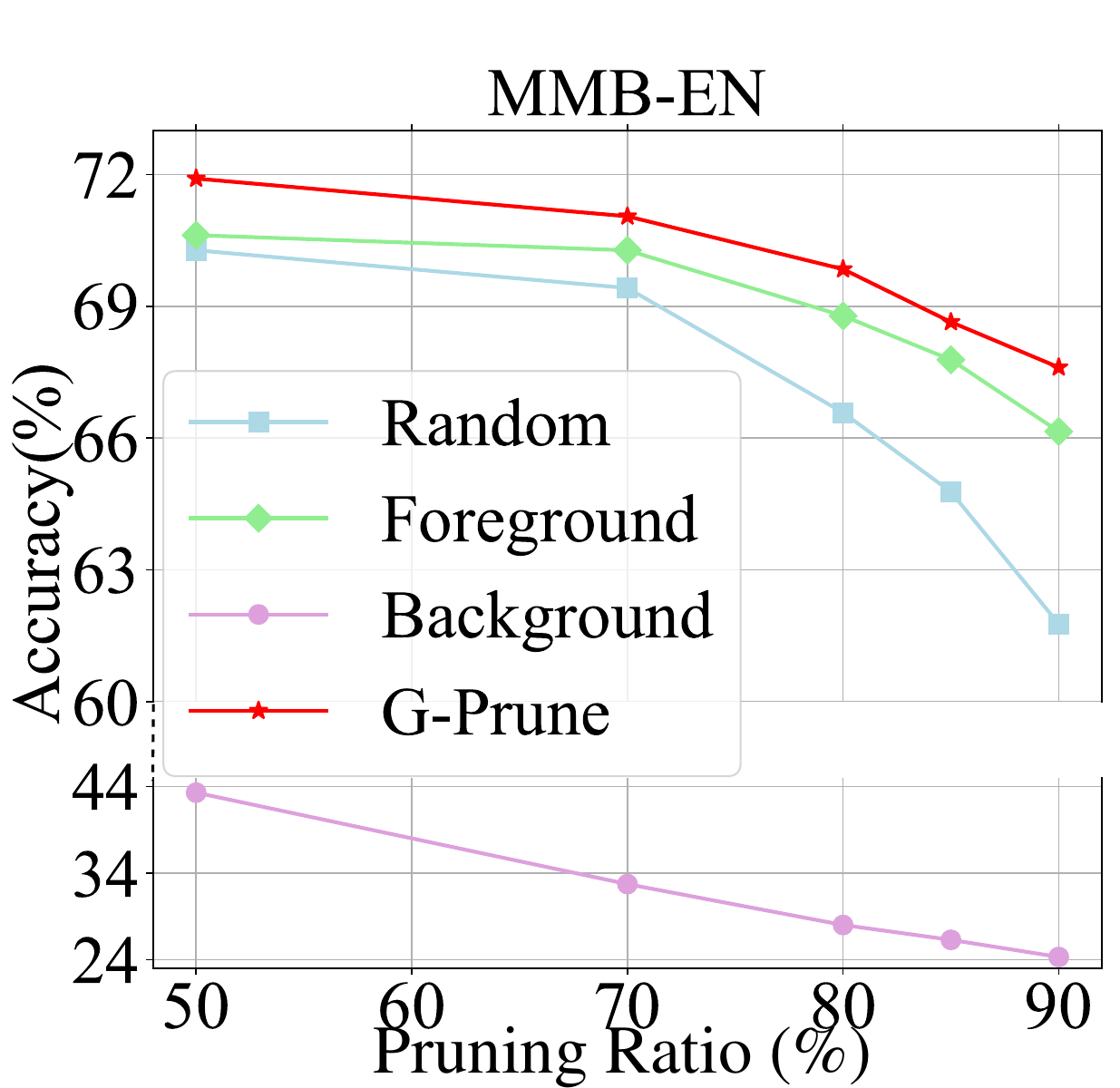}
    \end{minipage}%
    \begin{minipage}[c]{0.45\columnwidth} 
        \includegraphics[width=\linewidth]{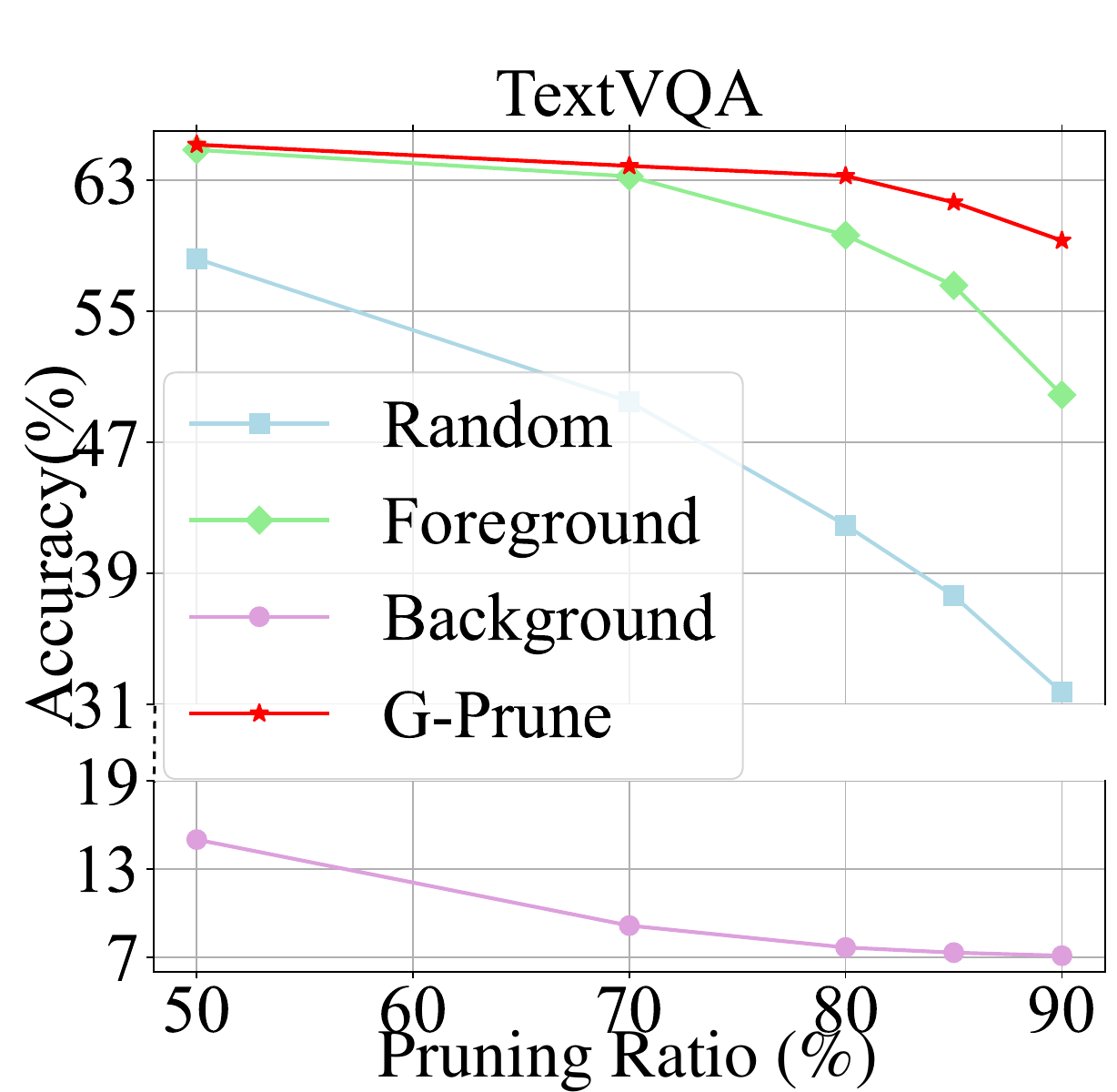}
    \end{minipage}
    
    \centering
    \begin{minipage}[c]{0.05\columnwidth} 
        \caption*{(b)} 
        \label{fig:b}
    \end{minipage}%
    \begin{minipage}[c]{0.9\columnwidth} 
        \includegraphics[width=\linewidth]{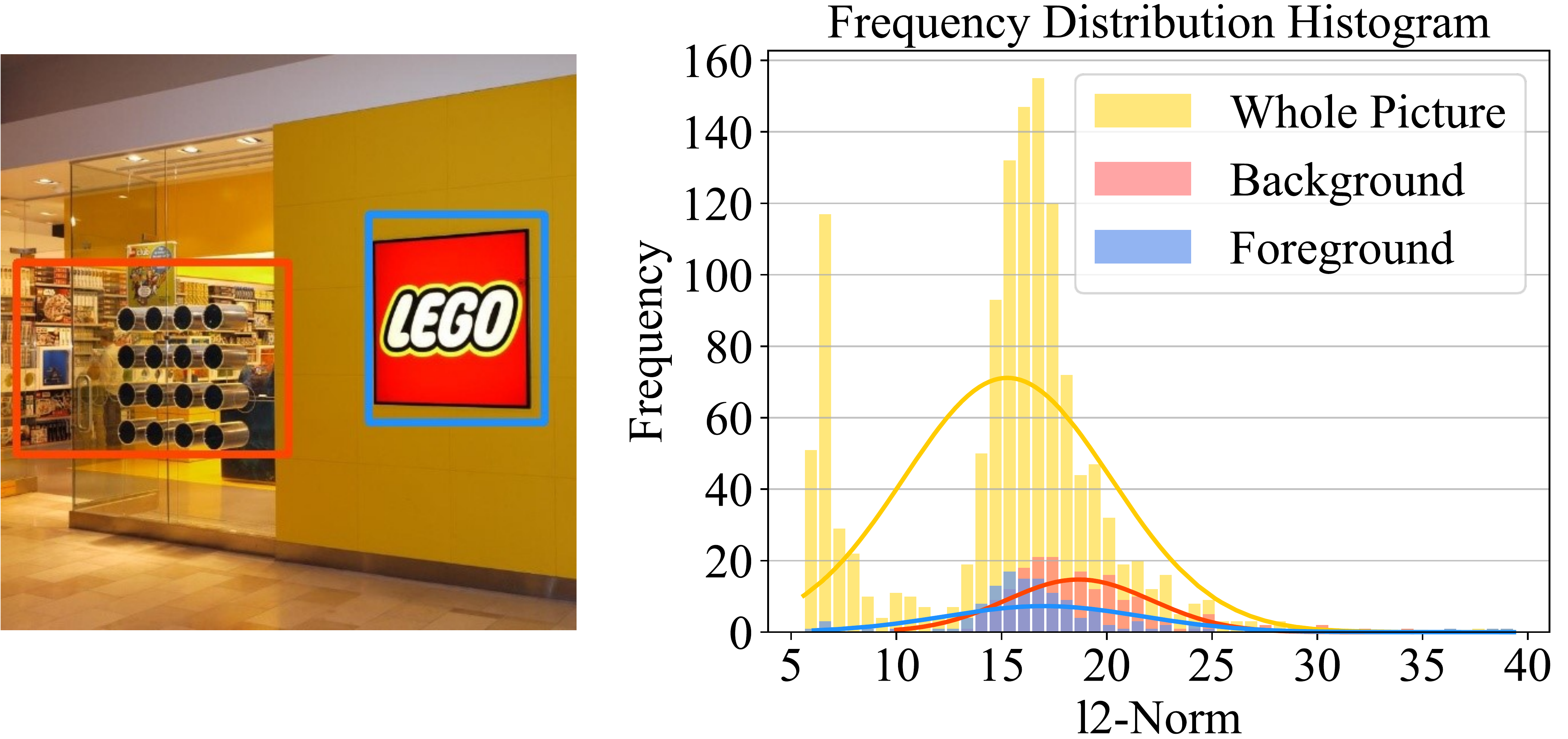}
    \end{minipage}

    \caption{(a) Random pruning, Foreground-preserving pruning, Background-preserving pruning and G-Prune performance curves in MMBench and TextVQA, (b) The frequency distribution of $l2$-Norm for the entire image, as well as its specific background and foreground areas. }
    \label{fig:Motivation}
\vspace{-5mm}
\end{figure}

In this case, a question rises, \emph{do we really need so many visual tokens for MLLMs}?
The large image tokens can provide more detailed visual semantics for multi-modal reasoning. However, these visual tokens are often processed by the well pre-trained image encoders, \emph{e.g.}, ViT \cite{dosovitskiy2020image}, and have a large receptive field, \emph{i.e.}, representing much more information than its actual area. In this case, heavy redundancy does exist in these image tokens \cite{pan2021ia}. Thus, randomly dropping a certain number of visual tokens has almost no impact on the performance on MMbench ~\cite{goyal2017making}, a widely used MLLM benchmark for common scenarios, also shown in Fig.\ref{fig:Motivation}-a. However, on the granular VL task, \emph{e.g.} TextVQA \cite{singh2019towards}, randomly removing a larger number of visual tokens will lead to a drastic decline in performance, indicating the great loss of key information, as shown in Fig.\ref{fig:Motivation}-a. Conclusively, the large number of visual tokens are apparently redundant, but what visual tokens to preserve remains an challenge given the varying difficulties of examples.

\begin{figure*}[t]
\begin{center}
\includegraphics[width=1.0\textwidth]{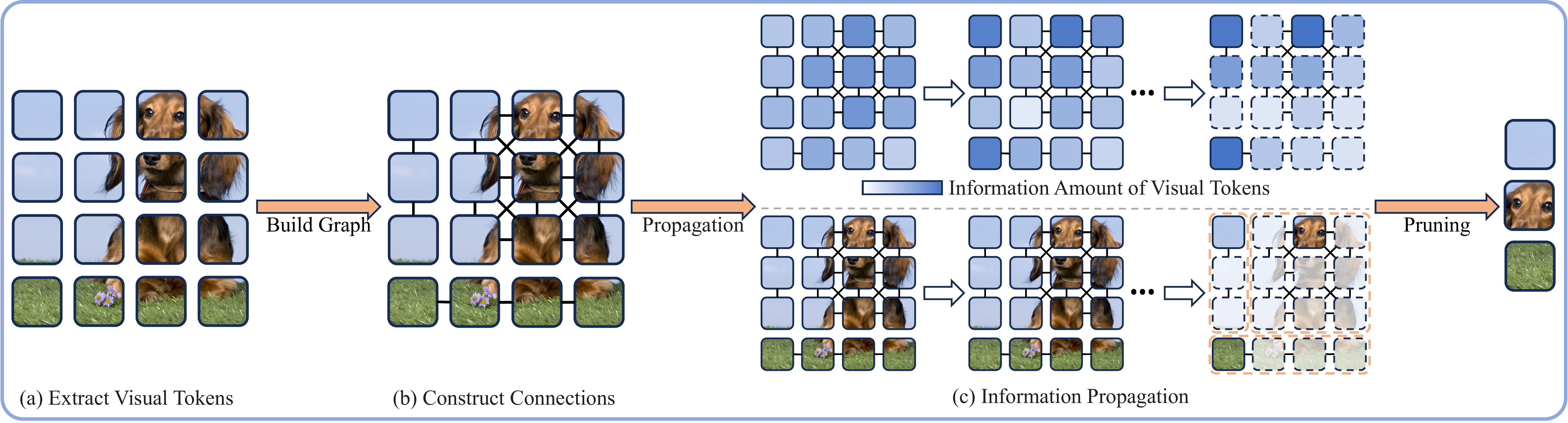}
\caption{
The overview of G-Prune.
G-Prune aims to find out the important visual tokens for MLLMs, thereby reducing the computation complexity. In practice, G-Prune regards all visual tokens as graph nodes, and build their connections based on their semantic similarities. Afterwards, an iterative algorithm is performed to propagate information among nodes and upgrade the importance scores of visual tokens. After iterations, we can select the top-$k$ tokens for MLLMs, which could be both foreground and background ones.
}
\label{fig:framework}
\vspace{-6mm}
\end{center}
\end{figure*}

Previous research \cite{rao2021dynamicvit,liang2022not,xu2022evo,yang2022focal} often focuses on the foreground information of the given images, \emph{i.e.}, the main objects, but we note that both the foreground and background tokens are important for MLLMs. To explain, for a conventional image recognition task, the foreground object mainly responds to the label of this image, and only a small number of foreground tokens can be capable of recognition. In contrast, for MLLM tasks, some details are often questioned. Meanwhile, we computed the $l2$-Norm frequency distribution histograms for both the foreground and background, and found that their distributions have significant overlap, as shown in Fig.\ref{fig:Motivation}-b. Nevertheless, how to recognize the important tokens for different images is still challenging in practice.

In this paper, we propose a graph-based method towards the training-free visual token pruning of MLLMs, termed G-Prune. In particular, we consider the visual tokens as graph nodes, and construct their connections according to their feature distances. Afterwards, information propagation is executed among nodes with an iterative algorithm to update the importance scores. Lastly, the most important tokens can be selected for MLLMs, which could be either foreground or background ones. In this way, we can select the most representative visual tokens for MLLMs, thereby greatly reducing the sequence length and computation complexity.

To validate G-Prune, we apply it to LLaVA-NeXT\cite{liu2024LLaVA}, and conduct extensive experiments on eight competitive MLLM benchmarks, including VQA2.0 \cite{goyal2017making}, GQA \cite{hudson2019gqa}, ChartQA~\cite{masry2022chartqa}, TextVQA \cite{singh2019towards}, DocVQA \cite{Mathew2020DocVQAAD}, POPE \cite{li2023evaluating}, MME \cite{fu2023mme} and MMB~\cite{liu2023mmbench}. The experimental results show that G-Prune significantly reduces computational costs while retaining high performance across various bench-marks. For instance, G-Prune can reduce the computation of LLaVA-NeXT on MME by 63.50\% without performance drops. Moreover, even on granular tasks like TextVQA, G-Prune can also achieve a higher accuracy than the compared pruning methods while dropping a larger number of tokens, e.g., 59.31 v.s. 39.02 (ToMe) on TextVQA while dropping 90\% tokens.

Conclusively, the contribution of this paper is three-fold:
\begin{itemize}
    \item We investigate the importance of different types of visual tokens for MLLMs, and show that both foreground and background ones are critical.
    
    \item We propose a graph-based method towards the training-free visual token pruning of MLLMs, which can mine the significant tokens via iterative information propagation.
    
    \item G-Prune can reduce the visual tokens of MLLMs to a large extent while retaining high performance on different VL benchmarks, even on the text-oriented benchmarks, such as TextVQA and DocVQA.
\end{itemize}

\section{Related Work}
\noindent\textbf{Multimodal Large Language Models}.
Driven by the remarkable success of \emph{large language models} (LLMs)\cite{touvron2023llama, touvron2023llama2, meta2024introducing, team2023gemini, reid2024gemini, achiam2023gpt, brown2020language} in managing text-only tasks with demonstrated exceptional capabilities, the field of \emph{multimodal large language models} (MLLMs)\cite{team2023gemini,achiam2023gpt,liu2024visual,liu2024LLaVA,luo2024cheap,internlmxcomposer2_5} is also attracting increasing scholarly interest.
The core principle underlying the extension of LLMs to MLLMs entails transforming visual information into a sequence of tokens. These tokens are then amalgamated with textual tokens to create a unified input for processing by LLMs.
Some approaches employ a series of learnable tokens to dynamically aggregate information from image sequences, which are then amalgamated with text tokens for processing within LLMs.
For instance, BLIP-2~\cite{li2023blip} introduces the QFormer module to dynamically schedule information aggregation from the outputs of the visual backbone.
Similarly, Qwen-VL~\cite{bai2023qwenvl} employs cross-attention mechanism to optimize the processing of information from ViT outputs.
Although these methods have demonstrated success and incur only limited additional computational overhead, the inadvertent loss of visual information during the process limits the upper limit of such methods~\cite{yao2024deco}.
To achieve this, other methods map visual information into the feature space of LLMs using a simple Multi-Layer Perceptron (MLP) layer.
For example, MINI-GPT4\cite{zhu2023minigpt} applies a projection layer to map visual features into the semantic space of the LLM.
Similarly, LLaVA~\cite{liu2024visual} follows this paradigm but further enhances it with a meticulously devised training strategy.
For higher resolution, LLaVA-NeXT~\cite{liu2024LLaVA} partitions images into multiple regions, and ViT is used to obtain visual features from upsampled images.
InternVL~\cite{chen2024internvl} presets a variety of high-resolution image inputs with different aspect ratios, and ViT is used to extract features from images of a specific size.
These methods effectively preserve visual information by fully retaining the visual tokens. 
While a large amount of redundant information is fed into LLMs.
To this end, it is necessary to filter out redundant tokens efficiently and effectively for the applications of MLLMs. 

\noindent \textbf{Token Pruning}.
Token pruning is a widely applied method for Transformer-based networks ~\cite{vaswani2017attention, devlin2018bert, dosovitskiy2020image}, which reduces the less important tokens to speed up inference.
In the aspect of MLLM, existing methods mainly use the calculation process of MLLM to locate redundant visual content.
For instance, FastV~\cite{chen2024image} uses the first two layers as the key to multimodal information exchange and directly removes most of the visual tokens with lower attention weights after the second layer.
Similarly, VTW~\cite{lin2024boosting} direct remove all visual tokens after a certain layer.
Some research efforts have also focused on pruning both textual and visual tokens simultaneously.
For instance, PuMer~\cite{cao2023pumer} reduces both textual and visual tokens in vision-language models by applying text-informed image pruning and modality-aware token merging. 
Moreover, MADTP~\cite{cao2024madtp} leverages the Multi-modality Alignment Guidance (MAG) module to align image and text features, and the Dynamic Token Pruning (DTP) module to adaptively prune tokens in both modalities at each layer.
The computational process involving MLLM will inevitably generate a lot of computational overhead.
To this end, directly adopting the pruning method for ViT may be an effective approach.
With additional module, DynamicViT~\cite{rao2021dynamicvit} reduces the number of tokens according to the probabilities generated by a small MLP layers. 
Besides, some training-free methods, \emph{e.g.}, EViT~\cite{liang2022not} and Evo-ViT~\cite{xu2022evo} calculate the similarity between the CLS token and other tokens, and remove or merge tokens with low similarity.
BAT~\cite{long2023beyond} combines token importance with matching and clustering techniques to effectively preserve both the most discriminative and diverse tokens.
These methods can effectively help us reduce computational overhead in ViT.
However, the results of highlighting the foreground while ignoring the background make them possibly not directly applicable to MLLM.
More deeply bound to the transformer computation process, Zero-TPrune~\cite{wang2024zero} initializes the importance of each token and continuously updates the importance of each token based on the attention matrix during the calculation process, and only the tokens with higher importance are retained.
Despite effectiveness, some existing methods take foreground as the main consideration ignores the requirement of the background in MLLMs. 
Others need to be integrated into the calculation process of MLLM, of which trials are expensive for MLLMs.
In this paper, we focus on a graph-based training-free method to filter out the redundant tokens while maintaining all objects by determining the most representative token of each object. 

\section{Method}

\begin{table*}[t]
\centering
\label{tab:main_table_llava}
\caption{
Comparison with SOTA methods under different FLOPs ratios for three strategies of benchmarks. The best results for each FLOPs ratios are marked in \textbf{bold}.
}
\resizebox{\textwidth}{!}{%
\begin{tabular}{l|c|c|cc|ccc|ccc|c}
\toprule
\multirow{2}{*}{Model} & \multirow{2}{*}{Method} & \multirow{2}{*}{\makecell{Pruning \\ Ratio}} & \multicolumn{2}{c|}{General VQA} & \multicolumn{3}{c|}{MLLM benchmarks} & \multicolumn{3}{c|}{Text-oriented VQA} & \multirow{2}{*}{\makecell{Throughput \\ (samples/sec)}}  \\
\cmidrule(lr){4-5} \cmidrule(lr){6-8} \cmidrule(lr){9-11}
             &        &       & GQA & VQA2.0 & MME & POPE & MMB-EN & ChartQA & DocVQA & TextVQA \\
\midrule
             & Baseline & 0\% & 65.38 & 82.70 & 1587.72 & 87.84 & 72.08 & 69.28 & 78.22 & 65.41 & 2.49\\
 \cmidrule(lr){2-12}
             &        &  50\% & 64.95 & 81.61 & 1605.43 & 86.47 & 70.27 & 56.00 & 61.54 & 58.21 & 4.03(1.62×)\\
             & Random &  70\% & 64.22 & 80.17 & 1576.33 & 84.98 & 69.42 & 44.48 & 48.42 & 49.48 & 5.23(2.10×)\\
             &        &  90\% & 60.55 & 74.23 & 1475.07 & 79.63 & 61.77 & 26.72 & 27.59 & 31.73 & 6.92(2.78×)\\
 \cmidrule(lr){2-12}
             &        &  50\% & 65.07 & 81.82 & 1566.60 & 87.56 & 70.88 & \textbf{68.20} & 73.53 & 59.07 & 1.00(0.40×) \\
             &  ToMe  &  70\% & 64.07 & 80.56 & 1564.36 & 87.33 & 68.21 & 62.88          & 65.28 & 52.19 & 1.11(0.45×) \\

LLaVA-NeXT-8B&        &  90\% & 59.72 & 76.31 & 1453.13 & 84.29 & 61.77 & 41.04          & 41.04 & 38.36 & 1.21(0.48×) \\
\cmidrule(lr){2-12}
             &        &  50\% & 65.11 & 82.51 & 1604.14 & 87.51 & 71.82 & 67.60 & 73.92 & 65.15 & 3.97(1.59×) \\
             &  FastV &  70\% & 64.34 & 81.83 & 1607.83 & 87.08 & 68.05 & 62.80 & 66.67 & 63.08 & 4.82(1.94×) \\
             &        &  90\% & 60.20 & 77.21 & \textbf{1488.16} & 83.01 & \textbf{68.30} & 39.56 & 42.57 & 53.53 & 5.90(2.37×) \\
\cmidrule(lr){2-12}
             &        &  50\% & \textbf{65.34} & \textbf{82.54} & \textbf{1623.81} & \textbf{87.76} & \textbf{71.91} & 67.72 & \textbf{75.62} & \textbf{65.17} & \textbf{4.03(1.62×)} \\
             & G-Prune(Ours)&  70\% & \textbf{64.37} & \textbf{81.91} & \textbf{1609.87} & \textbf{87.69} & \textbf{70.19} & \textbf{65.04} & \textbf{70.72} & \textbf{63.87} & \textbf{5.21(2.09×)}\\
             &        &  90\% & \textbf{61.40} & \textbf{77.51} & 1456.14 & \textbf{84.49} & 67.27 & \textbf{51.72} & \textbf{48.94} & \textbf{59.31} & \textbf{6.91(2.77×)} \\
\bottomrule
\end{tabular}%
}
\label{Table:Main-result}
\end{table*}
\begin{figure*}[t]
\begin{center}
\includegraphics[width=1.0\textwidth]{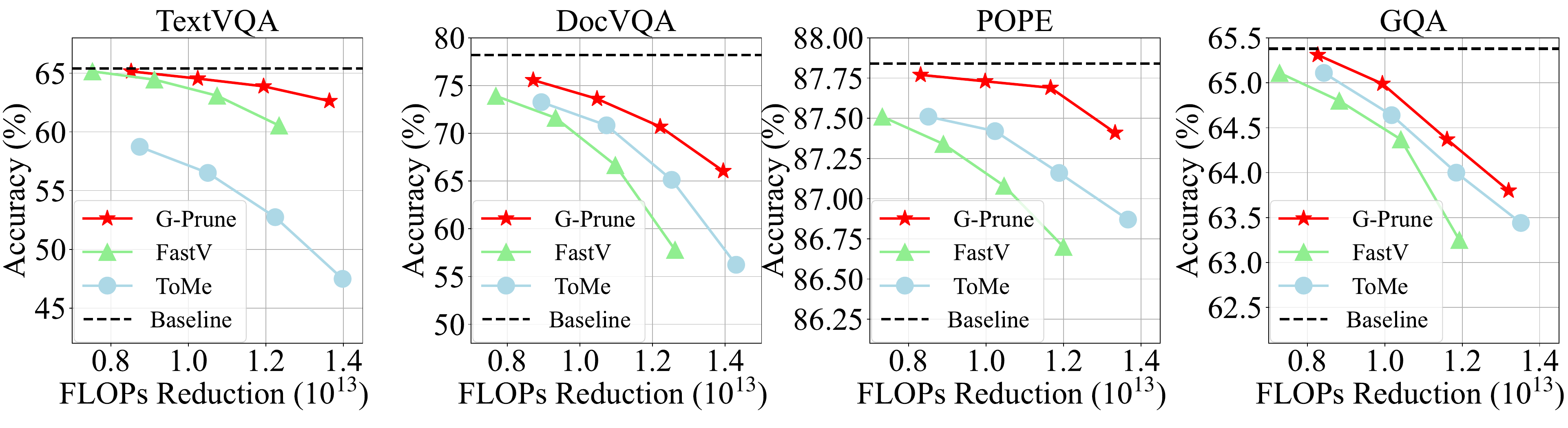}
\end{center}
\caption{
Comparison between our G-Prune and other compression methods for the LLaVA-NeXT model tested on the TextVQA, DocVQA, POPE and GQA benchmarks. 
}
\label{fig:FLOPs_Reduction_vs_Acc}
\end{figure*}

In this paper, we propose a graph-based method towards the training-free visual token pruning of MLLMs, termed \emph{G-Prune}. The principle of G-Prune is to select the most representative tokens from the perspective of graph, which could be either foreground or background ones. 

In practice, G-Prune considers the visual tokens as graph nodes and builds their connections based on the feature similarity. Afterwards, an iterative algorithm is executed to propagate information among nodes, based on which the most important ones emerge.

\begin{algorithm}[t]
\caption{G-Prune: Token Reduction via Graph-based Information Propagation}
\textbf{Input:} Visual tokens $\mathbf{X} \in \mathbb{R}^{N \times d}$, Similarity threshold $s$, Number of tokens to retain $k$, Number of iterations $t$

\textbf{Output:} Indices of selected representative tokens $I_k$
\begin{algorithmic}[1]
\State Construct $\mathbf{A} \in \mathbb{R}^{N \times N}$ according to Eq. \ref{Eqa:Similarity threshold}:
\For{$i, j = 1$ \textbf{to} $N$}
    \State $\mathbf{A}_{ij} \gets \max(0, \text{cos}(\mathbf{X}_i, \mathbf{X}_j) - s)$
\EndFor

\State Iterate information propagation:
\For{$t = 1$ \textbf{to} $t$}  
    \State $\mathbf{S}^{(t)} \gets \mathbf{S}^{(t-1)} \cdot (\mathbf{A}')$
\EndFor

\State Calculate degree of each token $\mathbf{D} \in \mathbb{R}^N$:
\For{$i = 1$ \textbf{to} $N$}
    \State $\mathbf{D}_i \gets \sum_{j=1}^N \mathbf{1}(\mathbf{A}_{ij} > 0)$
\EndFor

\State Normalize token scores $\mathbf{S}'^{(t)} \in \mathbb{R}^N$:
\For{$i = 1$ \textbf{to} $N$}
    \State $\mathbf{S}'^{(t)}_i \gets \mathbf{S}^{(t)}_i / \mathbf{D}_i$
\EndFor

\State Select top-$k$ representative tokens:
\State $I_k \gets \arg\max_{I \subset \{1,2,...,N\} | |I|=k} \sum_{i \in I} \mathbf{S}'^{(t)}_i$

\State \textbf{return} $I_k$
\end{algorithmic}
\label{alg:G-Prune}
\end{algorithm}

Concretely, given a set of visual tokens $\mathbf{X} \in\mathbb{R}^{N \times d}$, where $N$ and $d$ represent the number and dimension of visual tokens, we first construct the graph according to node-wise feature distances:
\begin{equation}
    \mathbf{A}_{ij} = 
    \begin{cases}
        cos(\textbf{X}_i, \textbf{X}_j), & cos(\textbf{X}_i, \textbf{X}_j) \geq s,\\
        0, &\text{otherwise},
    \end{cases}
    \label{Eqa:Similarity threshold}
\end{equation}
where $cos(\cdot, \cdot)$ represents the \emph{cosine} similarity between two tokens.
$s$ is a threshold for the connection. 
Then we apply the softmax function to normalize each row of \textbf{A}.
In this way, the tokens with similar semantics are connected, \emph{e.g.}, the background ones or the ones of the same object.

Afterwards, we iterate information propagation to find out the most representative token for each connected component. 
We first initialize the amount of information of visual tokens according to their $l2$-Norm:
\begin{equation}
\begin{aligned}
    \mathbf{S}^{(0)}_i &= \sqrt{\sum_{j=1}^d {\mathbf{X}_{ij}}^2}, \\
\end{aligned}
\label{con:L2-Norm}
\end{equation}
where $\mathbf{X}_{ij}$ denotes the $j$-th component of the $i$-th visual token. And $\mathbf{S}\in\mathbb{R} ^ { 1 \times N}$ denotes the initial score for each token. 
Then, the score of each token at the $t$-th step is obtained by 
\begin{equation}
    \mathbf{S}^{(t)} = \mathbf{S}^{(0)}(\mathbf{A}')^t.
    \label{Eqa:Iteration}
\end{equation}
Each object has a token that best represents it, and we gather scores for all tokens belonging to that object. However, since objects contain different numbers of tokens, their representative tokens can vary significantly. This variation makes it challenging to directly identify the most representative token.
To mitigate the influence of connected components size, we apply the degree of each token to normalize the scores.
The degree $\mathbf{D} \in \mathbb{R}^{N}$ of tokens can be calculated by
\begin{equation}
    \mathbf{D}_i = \sum_{j=1}^N\mathbf{1}(A_{ij} > 0),
\end{equation}
where $\mathbf{1}(\cdot)$ indicates an indicator function, which returns 1 when the condition is met.
Subsequently, the normalized score of each token can be defined as
\begin{equation}
    \mathbf{S}'^{(t)}_i =  \mathbf{S}^{(t)}_i / \mathbf{D}_i.
\end{equation}
Here, $\mathbf{S}'^{(t)}_i$ is the normalized representativeness of each token in its object.
Thanks to the normalization operation, we can use a unified standard to directly obtain the most representative token of each object.
Finally, the index of the retained tokens can be selected by
\begin{equation}
    I_k = \arg\max_{I \subset \{1,2,...,n\} | |I|=k} \sum_{i\in I} \mathbf{S}'^{(t)}_i.
\end{equation}
We find the top $k$ tokens by token score $\mathbf{S}'^{(t)}_i$.
In this way, we can retain the objects from both the foreground and the background, while only the most representative token is retained to represent each object.
The detailed algorithm of G-Prune is described in Alg.~\ref{alg:G-Prune}.

\section{Experiments}

\begin{table}[t]
\centering
\caption{
Ablation study of each component in G-Prune.
}

\setlength{\tabcolsep}{2pt} 
\begin{tabular}{cc|cccc|c}
\toprule
  $l2$-Norm  & Graph     & GQA   & POPE  & TextVQA & Avg. & Avg. TFLOPs \\ 
\midrule 
\ding{56}  & \ding{56}   & 64.22       & 84.98          & 49.48          & 66.23 & 6.76\\
\ding{52}  & \ding{56}   & 64.30       & 87.48          & 53.63          & 68.47 & 6.76\\
\ding{56}  & \ding{52}   & 64.23       & 87.46          & 64.01          & 71.90 & 6.76\\
\ding{52}  & \ding{52}   & \textbf{64.37}  & \textbf{87.69} & \textbf{64.05} & \textbf{72.04} & 6.76\\
\midrule
\multicolumn{2}{c|}{Baseline} & 65.38       & 87.84          & 65.41          & 72.88  & 18.52\\
\bottomrule
\end{tabular}
\label{table:modules}
\end{table}

\subsection{Datasets and Metrics}

\begin{figure}[t]
\begin{center}
\includegraphics[width=\columnwidth]{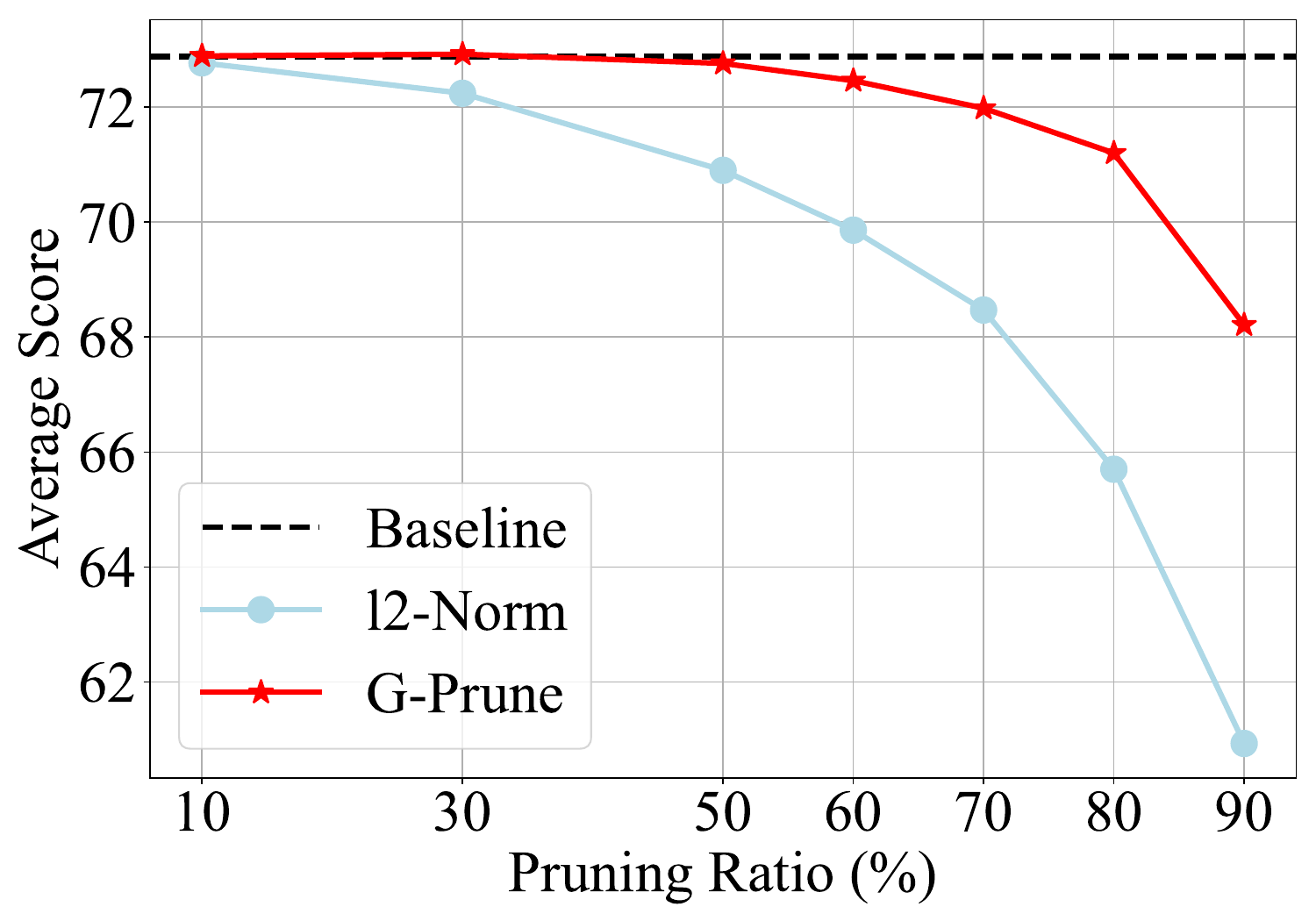}
\end{center}
\caption{
Comparison between our G-Prune method and $l2$-Norm based pruning method. The result is based on the average performance across GQA, POPE and TextVQA.
}
\label{Ablation_Keep_Ratio}
\end{figure}

We first evaluate G-Prune for two common general benchmarks, \emph{i.e.}, VQA2.0~\cite{goyal2017making}, GQA~\cite{hudson2019gqa}.
Furthermore, for fine-grained tasks, we evaluate OncePrue for text-oriented VQA benchmarks, including TextVQA~\cite{singh2019towards}, DocVQA~\cite{Mathew2020DocVQAAD} and ChartQA~\cite{masry2022chartqa}.
We also evaluate G-Prune on three emerging multimodal benchmarks for MLLMs, including POPE~\cite{li2023evaluating}, MME~\cite{fu2023mme} and MMB~\cite{liu2023mmbench}.
These benchmarks tend to be more challenging than traditional vision-language evaluations, aiming to assess different aspects of MLLMs, like detailed reasoning and OCR.

\subsection{Implementation Details}
We introduce G-Prune, a training-free, plug-and-play method that seamlessly integrates into existing MLLMs.
We implement G-Prune on widely-used open-source MLLM, LLaVA-NeXT-8B~\cite{liu2024LLaVA} following its default settings.
We use the evaluation toolkit LMMs-Eval~\cite{li2024lmms} to evaluate the performance of these MLLMs across various datasets. 
For specific implementations, we configure 5 iterations with a similarity threshold of 0.5.

\subsection{Quantitative Experiments}

\subsubsection{Comparison with state-of-the-art methods.}
In Tab.~\ref{Table:Main-result}, we compare the effect of G-Prune with existing pruning methods on LLaVA-NeXT, including ToMe ~\cite{bolya2022token}, FastV ~\cite{chen2024image}, and the random pruning.
From the Tab.~\ref{Table:Main-result}, we can first observe that, overall, our method achieves significant advantages in the above three types of data sets. 
On the general VQA tasks, the proposed G-Prune method has significant advantages at all pruning ratios, \emph{e.g.}, G-Prune improves performance by 0.88\%-1.58\% for VQA2.0 compared to ToMe. 
For the MLLM benchmarks, the proposed G-Prune improves performance on MME by 2.27\% while using only 50\% of the visual tokens. On the POPE dataset, our method consistently outperforms the baseline across all pruning ratios.
As for text-oriented VQA tasks, the proposed G-Prune can better maintain the performance compared to other methods.
When the pruning ratio is increased to 90\%, we can observe that the performance of previous methods, \emph{e.g.}, ToMe, has a significant decline, especially on text-oriented VQA benchmarks, \emph{i.e.}, the performance drops by 40.35\% on TextVQA and 26.02\% on ChartVQA.
In terms of ToMe, its complex indexing leads to large latency in MLLMs.
In contrast, the proposed G-Prune can better maintain the performance, \emph{e.g.}, G-Prune improves the performance by 51.99\% for TextVQA benchmark compared to ToMe when pruning 90\% tokens.
These results well validate the effectiveness of our G-Prune method in pruning visual tokens in the MLLM.

\begin{table}[t]
\centering
\caption{
Ablation study on the impact of hyper-parameters: Iteration num $t$ and Similarity threshold $s$.
}
\setlength{\tabcolsep}{3pt} 
\begin{tabular}{cc|cccc}
\toprule
Iteration $t$ & Similarity $s$ & GQA & POPE & TextVQA & Avg. \\ 
\midrule 
5 & 0.3 & 64.29 & 87.48 & 63.61 & 71.79 \\
5 & 0.5 & \textbf{64.37} & 87.69 & 63.87 & 71.98 \\
5 & 0.7 & 64.04 & \textbf{87.78} & \textbf{64.14} & \textbf{71.99} \\
5 & 0.9 & 63.87 & 87.22 & 54.20 & 68.43 \\
 \cmidrule(lr){1-6}
1 & 0.5 & 64.26 & \textbf{87.73} & 63.84 & 71.94 \\
5 & 0.5 & \textbf{64.37} & 87.69 & 63.87 & \textbf{71.98} \\
10 & 0.5 & \textbf{64.37} & 87.60 & \textbf{63.90} & 71.96 \\
50 & 0.5 & 64.05 & 87.34 & 63.85 & 71.75 \\
\bottomrule
\end{tabular}
\label{table:parameters}
\end{table}

\begin{figure*}[t]
\begin{center}
\includegraphics[width=1.0\textwidth]{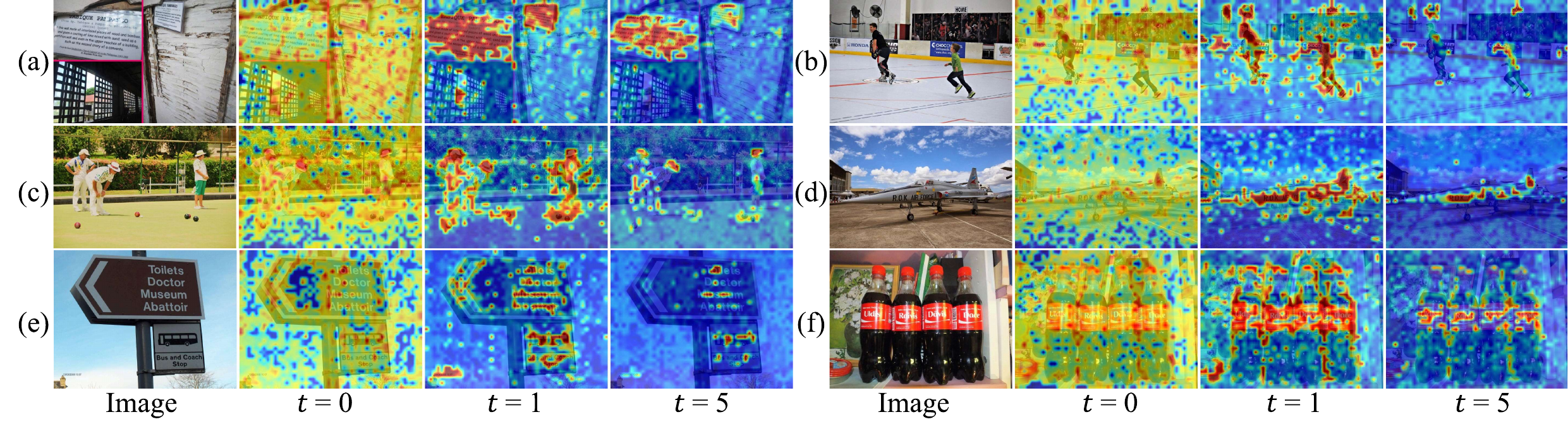}
\end{center}
\caption{
The visualization of information propagation in G-Prune across different iteration num $t$. The heatmaps demonstrates the score of each visual token on LLaVA-NeXT for TextVQA.
}
\vspace{-0mm}
\label{fig:Iteration}
\end{figure*}

For a more fine-grained comparison, we plot the computational cost-performance curve in Fig.~\ref{fig:FLOPs_Reduction_vs_Acc}.
From the curve, we can see that in the initial stage of reducing computational overhead, each method has no significant impact on the performance of the model on all benchmarks.
As the more FLOPs are reduced, the performance of the model begins to decline rapidly.
Specifically, on POPE benchmark, when reducing FLOPs from \(1.83 \times 10^{13}\) to \(6.69 \times 10^{12}\), there is only 0.09\% performance decreased.
While reducing FLOPs from \(1.83 \times 10^{13}\) to \(3.37 \times 10^{12}\), the performance significantly drops by 4.11\%.
This phenomenon shows that there is obvious redundancy in the visual tokens of MLLM. 
Therefore, in the initial stage, all methods can achieve token pruning without performance loss.
However, as the number of pruning increases, tokens must be selected more carefully to achieve competitive performance.
Meanwhile, we can observe that the proposed G-Prune method maintains the great advantages over other methods under different FLOPs reductions.
For instance, when reducing \(1.53 \times 10^{13}\) FLOPs for TextVQA benchmark, the proposed G-Prune improves performance by 52.05\% compared to ToMe.
As for reducing \(1.41 \times 10^{13}\) FLOPs, the proposed G-Prune keep 46.67\% advantage than FastV for DocVQA benchmark.
These results better validating the advantages of G-Prune for token pruning task in MLLM.

\subsubsection{Comparison with Alternatives.}
In Fig.~\ref{Ablation_Keep_Ratio}, we compare our G-Prune with the alternative using $l2$-Norm as the metric for LLaVA-NeXT on GQA, POPE and TextVQA benchmarks.
The curve reflects the average performance of GQA, POPE and TextVQA under different pruning ratios. 
Specifically, \emph{baseline} denotes the default MLLM without token pruning. 
``$l2$-Norm'' indicates retaining the tokens with the largest $l2$-Norm value according to the percentage.

As shown in Fig.~\ref{Ablation_Keep_Ratio}, it is apparent that when the retention tokens exceed 50\%, G-Prune consistently maintains stable performance compared to the baseline method, \emph{i.e.}, G-Prune still retains 99.84\% of the original performance.
At the same time, ``$l2$-Norm'' method drops performance by 2.56\%.
Based on this, we can find that in the complex scenarios that MLLM needs to deal with, it is impossible to achieve token pruning by relying solely on the amount of information.
Moreover, across all evaluated pruning ratios, G-Prune performs better than the method according to $l2$-Norm only.
This advantage grows more pronounced as the pruning ratio increases.
For instance, when the pruning ratio is increased to 90\%, our G-Prune can improve the performance by 7.28\% compared to ``$l2$-Norm'' method.
It fully shows that it is effective to measure the value of tokens by constructing a similarity graph between tokens in the visual token pruning task.
These experiment results fully demonstrate that the proposed graph-based information propagation method in our G-Prune can effectively help us better perform visual token pruning in MLLMs.

\subsubsection{Ablation Study.}

Further, we conduct ablation experiments for each component in Tab.~\ref{table:modules} to explore the contributions and impacts of different components of our method.
From Tab.~\ref{table:modules}, we can first observe that when randomly pruning 30\% tokens, \emph{i.e.}, the first line in Tab.~\ref{table:modules}, there is a significant performance drop, \emph{e.g.}, the performance drops by 22.75\% for TextVQA dataset. 
Then, we apply $l2$-Norm to measure the value of each token, and directly prune the tokens with lower l2-norm, this approach primarily preserves elements with higher information content~\cite{lu2022norm}, resulting in an average performance of 68.47.
Conversely, we employ information propagation by constructing graph, and initial the score of each token to 1.
With the better consideration to objects from both foreground and background, this approach improves the performance by 8.56\% compared to $l2$-Norm only method.
Finally, when applying both two methods, \emph{i.e.}, our G-Prune method, the average performance is improved by 8.77\% compared to the random pruning method.
These experiment results not only validate the effectiveness of each component in our G-Prune, but also show that every object from the foreground and background should be fully considered during the token pruning of MLLM.

\begin{figure*}[t]
\begin{center}
\includegraphics[width=1.0\textwidth]{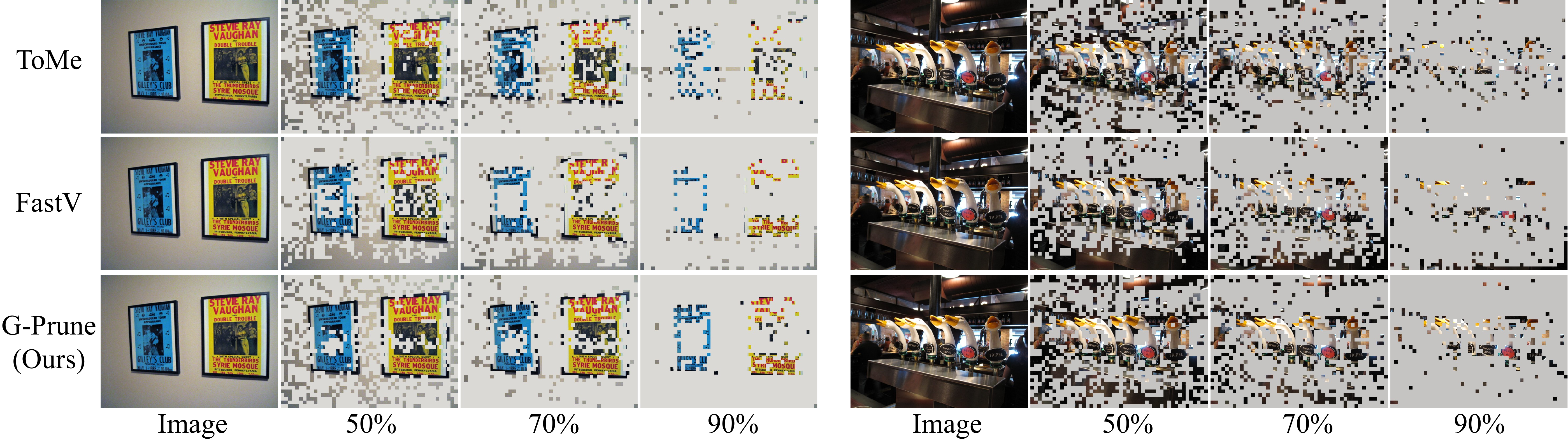}
\end{center}
\caption{
The comparative visualization of ToMe, FastV, and G-Prune on LLaVA-NeXT. G-Prune effectively retains tokens representative of regions with high information content, giving it a significant advantage in fine-grained tasks.
}
\vspace{-0mm}
\label{fig:Visualization}
\end{figure*}

In Tab.~\ref{table:parameters} we conduct experiment based on the different hyper-parameter settings.
We first fix the number of iteration turn $t$ in Eq.\ref{Eqa:Iteration}. 
When the similarity threshold $s$ in Eq.\ref{Eqa:Similarity threshold} gradually increased from 0.3 to 0.9, we can observe that the average performance gradually increase to 2.78\% when $s=0.7$.
With the further increment of the similarity threshold $s$, the performance is dropped by 4.95\%.
The reason for this phenomenon is that a too low similarity threshold $s$ will cause all regions to be regarded as the same object, while a too high s will cause an object to be split into multiple objects unnecessarily.
When fix the similarity threshold $s$ to 0.5, we can observe that the best performance occurred with $5$ iterations, \emph{i.e.}, the average performance is achieved by 71.98.
It is worth noting that the performance difference of our G-Prune is very limited under different iteration numbers. 
Specifically, the difference between the highest and the lowest is only 0.32\%.
It is because that the propagation of information will reach equilibrium after a certain number of rounds, \emph{i.e.}, the amount of information flowing in and out of each token is exactly the same.
To this end, the score of tokens will have no change.
The above experiment not only validate the effectiveness of our G-Prune method, but also shows that our G-Prune method has good robustness.

\subsection{Qualitative Experiments}
To further delve into the intrinsic mechanisms of our method, we visualize the information propagation in Fig.~\ref{fig:Iteration}, the heatmap represents the score of tokens. 
The tokens in the red areas have higher scores.
In the initial stage, multiple tokens have high scores for the same object. 
This also shows that there is a lot of redundant information. 
As the information continues to propagate, we can find that the information gradually gathers in several tokens from both foreground and background. 
This shows that our G-Prune can effectively retain representative tokens for objects.
These experiment results clearly prove the effectiveness of our G-Prune in visual token pruning.
Then we visualize the results on the LLaVA-NeXT using the TextVQA dataset, pruning 50\%, 70\%, and 90\% of the tokens to explore the pruning mechanisms of our method. We also compare our G-Prune method with ToMe and FastV.
As shown in Fig.~\ref{fig:Visualization}, we first focus on the performance of our method under different pruning ratios. When keep 50\% retain tokens, the foreground occupies a large proportion, while the background area is preserved uniformly. This phenomenon shows that our G-Prune can effectively preserve information for each area. With the decreasement of retain tokens, we can observe that more tokens will be retained in areas with complex textures, whether from the background or the foreground. At the same time, in each area with high similarity, a certain number of tokens will be retained. 
Subsequently, we conducted a visual comparison of our method with others. For instance, in the left image, even with a pruning ratio of 50\%, ToMe and FastV begin to lose fine-grained details, such as text on a wall image. This situation is even more serious at a 90\% pruning ratio. Our method not only preserves more detailed information but also significantly reduces redundancy by targeting areas with higher similarity.
On the other hand, from the right image, we can find out that the visual tokens from the background are completely pruned.
This may cause MLLM to be unable to solve some tasks involving background information.
Above all, with the decreasement of retain tokens, we can observe that more tokens will be retained in areas with complex textures, whether from the background or the foreground. 
At the same time, in each area with high similarity, a certain number of tokens will be retained. 
Under this paradigm, our G-Prune method can better preserve the information in the original image.
These visualizations prove that G-Prune not only effectively maintains representative visual tokens, but also well alleviates redundancy.
\section{Conclusion}

In this paper, we introduced a novel, training-free visual token pruning method for multimodal large language models (MLLMs) named G-Prune.
Our G-Prune method does not need to intervene in the reasoning process of MLLM, saving a lot of computational overhead. 
Meanwhile, the proposed G-Prune method considers objects in the foreground and background at the same time, fully meeting the complex problems that MLLM needs to deal with.
Specifically, our approach constructs a graph based on token similarities, iteratively propagating information to identify and retain the most representative tokens for objects from both foreground and background.
Extensive experiments across various benchmarks prove superiority of G-Prune over existing pruning methods, maintaining a great balance between computational efficiency and performance.
For instance, G-Prune achieves a 63.57\% FLOPs reduction on LLaVA-NeXT for TextVQA while only incurring a 2\% decrease in accuracy.
These findings suggest that G-Prune significantly advances the efficiency and scalability of MLLMs.

\section{Acknowledgments}

This work was supported by the National Science Fund for Distinguished Young Scholars (No.62025603), the National Natural Science Foundation of China (No. U21B2037, No. U22B2051, No. U23A20383, No. U21A20472, No. 62176222, No. 62176223, No. 62176226, No. 62072386, No. 62072387, No. 62072389, No. 62002305 and No. 62272401), and the Natural Science Foundation of Fujian Province of China (No. 2021J06003, No.2022J06001).

\bibliography{aaai25}

\end{document}